# Performance Evaluation of Convolutional Neural Networks for Gait Recognition

K.D. APOSTOLIDIS, P.S. AMANATIDIS, G.A. PAPAKOSTAS[*]

HUman-MAchines Interaction Laboratory (HUMAIN-Lab), Department of Computer Science, International Hellenic University, Kavala, 65404 Greece

In this paper, a performance evaluation of well-known deep learning models in gait recognition is presented. For this purpose, the transfer learning scheme is adopted to pre-trained models in order to fit the models to the CASIA-B dataset for solving a gait recognition task. In this context, 18 popular Convolutional Neural Networks (CNNs), were re-trained using Gait Energy Images (GEIs) of CASIA-B containing almost 14000 images of 124 classes under various conditions and their performance was studied in terms of accuracy. Moreover, the performance of the studied models is managed to be explained by examining the parts of the images being considered by the models towards providing their decisions. The experimental results are very promising, since almost all the models achieved a high accuracy over 90%, which is robust to the increasing number of classes. Furthermore, an important outcome of this study is the fact that a recognition problem can be effectively solved by using CNNs pre-trained to different problems, thus eliminating the need of customized model design.

## 1 INTRODUCTION

Biometrics is a specialized area of research, which analyses biological data through statistical and mathematical methods. Biometric features are mainly used for human identification because they are almost unchangeable over time. Some of the most well-known biometric features are fingerprint, iris, signature, face, voice and the geometry of hand. Gait recognition constitutes a biometric feature with various applications in completely different fields. Human identification using gait recognition represents several advantages over the other biometric features. The most important property is that it is quite discreet since it does not need to be in contact or close to the sensor or the camera and does not require high resolution data. At the same time, it is difficult to hide, to imitate or to pretend the biometric feature [1]. Nevertheless, it has been proven that there are some inhibitory factors, which complicate the proper implementation of gait recognition. One of the most important is the surface in which the person walks. For example, a wet surface affects the speed, the length of the gait and the angle of the sole [2]. Another significant factor is a possible injury, which can affect the control of the body. Moreover, the age, the clothing and the carriage of a bag can decrease the efficacy of the system [3]. Gait analysis is a systematic study of human movement. This type of analysis includes measuring, describing and evaluating the quantities that characterize a human movement [4].

---

[*] Corresponding author: gpapak@teiemt.gr

The contribution of this work is two-fold: (1) a performance evaluation of the most popular CNNs in gait recognition using GEIs [5] is conducted, for the first time and (2) an interpretation of the applied CNNs performance is presented. This work aims to investigate how a set of pre-trained CNNs can be used to solve a novel problem, after training with new data, without the need to design a customized model.

The rest of the paper is organized as follows: Section 2 summarizes the related work in the field of gate recognition with emphasis on the approaches that use deep learning models. Section 3 provides a brief discussion about the models used in this study, while Section 4 presents the results of the conducted simulations. Finally, Section 5 concludes this work.

## 2 RELATED WORK

The most used approach for the problem of gait recognition is deep learning and convolutional neural networks [6] as it is the most powerful tool in computer vision applications. A Siamese neural network for human identification through gait analysis was presented by Zhang et al. [7]. The specific network automatically extracts the robust and distinct features of a walk. GEIs (Gait Energy Images) is used for the most efficient training of the network in cases of limited data, instead of raw data from walking sequences, which are also used as inputs to the network. Alotaibi et al. [8] attempted to solve the problem of gait recognition using a CNN architecture, which is less sensitive to various variations or obstacles that may arise. At the same time, while deep learning models require large data sets, this particular architecture can manage small data sets without increasing them. The design has eight layers: four convolutional layers and four subsampling layers, where each layer has 8 feature maps. Furthermore, Wolf et al. [9] proposed a 3D CNN that uses space-time information to find a generalized description of human walking that would not depend on angle view, color, and different walking conditions. The network uses 3x3x3 filters (Height, Width, Temporal) as suggested in a previous paper [10], as this allows motion detection in all directions and includes future and previous time information, using the smallest possible filter. In 2017, Yu et al. [11] presented a paper, which uses Generative Adversarial Networks (GAN) to identify humans from walking. The presented method is called GaitGAN and its purpose is to create a GAN model that will produce images from the dataset CASIA-B [12], which will be unchanged in terms of angle view and independent of clothing and possible objects that may bring the subject. The biggest challenge in this methodology is that when creating these images, identification information must not be lost for the model to work properly. At the same time, GaitGAN differs from the usual GANs in terms of discriminators as it has two and not one. This results in improved authentication in addition to creating images. In 2016, Kohei et al. [13] proposed GEINet, which is a CNN with gait energy images as inputs. The GEINet consists of 8 layers and it is trained in the OU-ISIR [14] large population dataset.

## 3 MODEL DESCRIPTION

Herein, a set of 18 pre-trained CNN models is evaluated towards identifying a person by utilizing gait information. Table 1 summarizes the characteristics of the examined CNN models. The ResNet is a residual learning framework to ease the training of networks. This model reformulates the layers as learning residual functions with reference to the layer inputs, instead of learning unreferenced functions. The residual networks are easier to optimize and can gain accuracy from considerably increased depth. MobileNets are based on a streamlined architecture that uses depth-wise separable convolutions to build light weight deep neural networks and are used in embedded vision applications. Dense Convolutional Network (DenseNet) introduces direct connections between any two layers with the same feature-map size. DenseNets scale naturally to hundreds of layers while exhibiting no optimization difficulties. VGG16 and VGG19 are Deep convolutional networks with up to 19 weight layers. These models are showing the effect of the convolutional network depth on its accuracy in the large-scale image recognition. The Xception model has a similar parameter count as Inception V3. Compared to Inception V3, Xception shows small gains in classification performance on the ImageNet dataset and large gains on the JFT [15] dataset. All these models were trained on the ImageNet [16] dataset.

Table 1. The depth and parameters of pre-trained models

| Model | Parameters | Depth (#layers) |
|---|---|---|
| MobileNet [17] | 4.253.864 | 88 |
| ResNet50 [18] | 25,636,712 | 177 |
| InceptionV3 [19] | 23,851,784 | 159 |
| Xception [20] | 22,910,480 | 126 |
| DenseNet121 [21] | 8,062,504 | 121 |
| DenseNet169 [21] | 14,307,880 | 169 |
| DenseNet201 [21] | 20,242,984 | 201 |
| NASNet Mobile [22] | 5,326,716 | 771 |
| NASNet Large [22] | 88,949,818 | 1041 |
| InceptionResNet [23] | 55,873,736 | 572 |
| VGG16 [24] | 138,357,544 | 23 |
| VGG19 [24] | 143,667,240 | 26 |
| ResNet101 [18] | 44,707,176 | 347 |
| ResNet152 [18] | 60,419,944 | 517 |
| MobileNetV2 [25] | 3,538,984 | 88 |
| ResNet50V2 [18] | 25,613,800 | 192 |
| ResNet101V2 [18] | 44,675,560 | 379 |
| ResNet152V2 [18] | 60,380,648 | 566 |

## 4 EXPERIMENTAL STUDY

For the sake of the experiments the pre-trained versions of the models of Table 1, as implemented in Keras library [26]. All the experiments were carried out on a laptop computer equipped with Intel i7-6700HQ CPU, 8GB DDR4 RAM and GTX 960M GPU.

**4.1 Experiments Analysis**

We experimented with pre-trained models to find the best parameter tuning. We changed the epochs and the number of the trainable layers, so that each model offers the best possible result. Almost all models trained for 10 epochs except for NASNet Large that trained for 15 epochs. The models were re-trained with the GEI CASIA

B [12] dataset for 50, 75, 100 and 124 classes. The used evaluation metrics were *Accuracy (Acc.)*, *Recall (Rec.)* and *Precision (Pr.)*. In Table 2 the results are presented for each model for all metrics.

Table 2. Evaluation of models for 50 and 75 classes

| Model | Trainable Layers | Acc./ Rec. / Pr. 50 Classes (%) | | | Acc./ Rec. / Pr. 75 Classes (%) | | | Acc./ Rec. / Pr. 100 Classes (%) | | | Acc./ Rec. / Pr. 124 Classes (%) | | |
|---|---|---|---|---|---|---|---|---|---|---|---|---|---|
| MobileNet | 20 | **99.8** | **99.6** | **99.8** | 94.6 | 96.2 | 97.3 | 96.6 | 92.8 | 95.2 | 95.2 | 96.1 | 98.5 |
| ResNet50 | 10 | 97.6 | 94.8 | 98.5 | 92.5 | 91.3 | 92.6 | 94.6 | 88.7 | 95.1 | 92.5 | 88.7 | 93.7 |
| InceptionV3 | 30 | 95.6 | 93.4 | 98.0 | 88.5 | 87.7 | 94.2 | 91.1 | 87.6 | 94.2 | 85.0 | 81.0 | 94.9 |
| Xception | 20 | 98.0 | 96.2 | 94.8 | 91.5 | 91.0 | 94.6 | 94.0 | 89.6 | 95.8 | 94.1 | 92.5 | 95.1 |
| DenseNet121 | 40 | 98.2 | 93.0 | 99.1 | 92.6 | 90.0 | 96.5 | 94.9 | 88.4 | 97.8 | 93.6 | 87.6 | 98.3 |
| DenseNet169 | 30 | 98.6 | 93.0 | 99.3 | 92.5 | 86.7 | 97.1 | 95.1 | 89.1 | 97.8 | 95.1 | 87.6 | 97.3 |
| DenseNet201 | 40 | 99.4 | 95.2 | 99.8 | 94.6 | 91.6 | 97.9 | **97.7** | **93.0** | **98.4** | **95.6** | **92.6** | 98.3 |
| NASNet Mobile | 60 | 84.0 | 74.4 | 91.2 | 79.8 | 70.0 | 90.6 | 83.7 | 72.7 | 91.5 | 80.1 | 74.4 | 90.3 |
| NASNet Large | 40 | 88.6 | 86.7 | 84.3 | 80.2 | 73.4 | 84.3 | 77.2 | 75.1 | 79.1 | 75.8 | 73.4 | 81.2 |
| InceptionResNet | 20 | 94.4 | 89.0 | 92.6 | 86.2 | 83.7 | 92.7 | 86.1 | 83.6 | 94.1 | 89.2 | 88.2 | 91.8 |
| VGG16 | 20 | 97.0 | 97.0 | 96.5 | 90.3 | 92.2 | 88.5 | 92.2 | 92.0 | 95.4 | 92.5 | 90.0 | 94.6 |
| VGG19 | 10 | 92.6 | 92.7 | 98.1 | 90.3 | 90.6 | 91.2 | 94.6 | 90.2 | 94.4 | 91.5 | 90.0 | 92.7 |
| ResNet101 | 40 | 99.4 | 86.2 | 95.8 | 86.6 | 96.9 | 96.2 | 95.8 | 93.0 | 94.3 | 91.0 | 94.2 | 96.4 |
| ResNet152 | 30 | 97.8 | 95.2 | 98.5 | 93.8 | 83.6 | 88.7 | 97.2 | 95.1 | 95.7 | 87.2 | 94.8 | 95.4 |
| MobileNetV2 | 20 | 97.6 | 96.0 | 98.8 | 90.7 | 89.9 | 96.0 | 93.2 | 92.9 | 95.4 | 90.0 | 90.1 | 94.5 |
| ResNet50V2 | 30 | 99.0 | 99.2 | 97.3 | 93.6 | 94.4 | 94.3 | 96.9 | 90.7 | 96.2 | 94.6 | 94.0 | 96.4 |
| ResNet101V2 | 30 | 97,8 | 95.9 | 95.5 | 91.7 | 93.3 | 93.3 | 96.8 | 93.2 | 96.3 | 93.0 | 94.9 | 94.8 |
| ResNet152V2 | 40 | 97.4 | 98.0 | 97.7 | 92.6 | 95.3 | 93.0 | 93.1 | 93.5 | 95.1 | 93.0 | 95.7 | 94.9 |

The above results reveal that the performance of almost all the models decreases with the increase in the number of classes. However, DenseNet201 shows more robust to the increasing of the problem complexity followed by the MobileNet, However, considering the small size (Table 1) of the MobileNet compared to the DenseNet201 and the rest of the models, it is concluded that this is the best choice for gait recognition.

**4.2 Interpretability**

Although deep learning provides high accuracy in difficult recognition problems, its "black box" nature does not allow enough interpretability capabilities. Recently, scientists try to explain why the models work properly. A popular method in this direction is the Grad-CAM (Gradient- weighted Class Activation Mapping) [27], which shows where the model "look at" in the image to make a decision. This method is applied in this work for the top 5 models of the previous analysis for the case of three test images, and the produced maps along with the decision confidence in each case are presented in Table 3.

Table 3. Grad-CAM of top 5 models and their confidence for the case of the 124 classes

| Model | Image 1 | Image 2 | Image 3 | Confidence (%) Image 1 | Image 2 | Image 3 |
|---|---|---|---|---|---|---|
| DenseNet 201 | 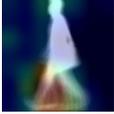 | 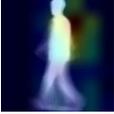 | 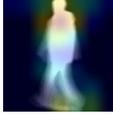 | 96.8 | 97.0 | 83.5 |
| DenseNet 169 | 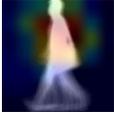 | 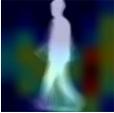 | 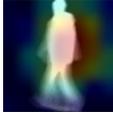 | 99.2 | 90.0 | 90.6 |
| MobileNet | 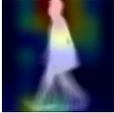 | 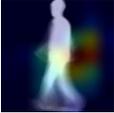 | 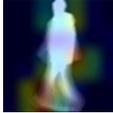 | 99.9 | 99.9 | 96.1 |
| Xception | 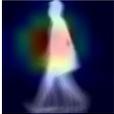 | 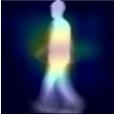 | 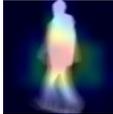 | 99.4 | 99.6 | 99.7 |
| ResNet 50V2 | 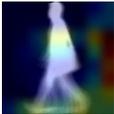 | 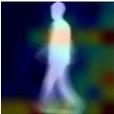 | 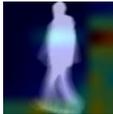 | 99.9 | 99.3 | 76.1 |

Two very important observations are derived from the results of Table 3: (1) the high accuracy of the models is highly related to the features extracting from the middle part of the human's body, (2) the legs are not the most informative part of the body that characterizes uniquely the gait pattern as it is expected.

## 5  CONCLUSION

This work studied the ability of 18 pre-trained CNN models to solve the problem of gait recognition. The experiments shown that the models can achieve high accuracy over 90%, for the case of the CASIA-B dataset, and thus transfer learning is able to adopt existing re-trained models to new applications with great success. More importantly, this study used the Grad-CAM method in order to interpret the decisions of the models by revealing that the middle part of human's body carries distinct information relative to gait biometrics. However, experiments with more data should be arranged in the future in order to derive more concrete conclusions.

## ACKNOWLEDGMENTS

This work was supported by the MPhil program "Advanced Technologies in Informatics and Computers", hosted by the Department of Computer Science, International Hellenic University, Greece.